\documentclass[conference]{IEEEtran}
\IEEEoverridecommandlockouts
\usepackage{cite}
\usepackage{amsmath,amssymb,amsfonts}
\usepackage{algorithmic}
\usepackage{graphicx}
\usepackage{textcomp}
\usepackage{xcolor}
\def\BibTeX{{\rm B\kern-.05em{\sc i\kern-.025em b}\kern-.08em
    T\kern-.1667em\lower.7ex\hbox{E}\kern-.125emX}}
\begin{document}

\title{Realistic face animation generation from videos}

\author{\IEEEauthorblockN{1\textsuperscript{st} Zihao Jian}
\IEEEauthorblockA{\textit{dept. Computing Science} \\
\textit{Multimedia}\\
\textit{University of Alberta}\\
Edmonton, Canada \\
zjian3@ualberta.ca}
\and
\IEEEauthorblockN{2\textsuperscript{nd} Minshan Xie}
\IEEEauthorblockA{\textit{dept. Computing Science} \\
\textit{Multimedia}\\
\textit{University of Alberta}\\
Edmonton, Canada \\
minshan1@ualberta.ca}

}

\maketitle

\begin{abstract}

3D face reconstruction and face alignment are two fundamental and highly related topics in computer vision. Recently, some works start to use deep learning models to estimate the 3DMM coefficients to reconstruct 3D face geometry. However, the performance is restricted due to the limitation of the pre-defined face templates. To address this problem, some end-to-end methods, which can completely bypass the calculation of 3DMM coefficients,
are proposed and attract much attention. In this report, we introduce and analyse three state-of-the-art methods in 3D face reconstruction and face alignment. Some potential improvement on PRN are proposed to further enhance its accuracy and speed.
\end{abstract}


\section{Introduction}
\label{intro}
3D face reconstruction has attracted much attention in the academic research field. Compared with 2D face images, 3D faces can provide much more space information from different angels. Therefore, using 3D face models as dataset in face recognition typically contributes to higher accuracy and better robustness than using 2D images. Besides, 3D face models are also widely utilized in many different fields such as animation\cite{animation}, image restoration\cite{restoration}, anti-spoofing\cite{anti-spoofing}, and tracking\cite{tracking}. However, it is still an open problem to apply 3D face models in these fields due to the scarce data of 3D faces with high resolution. Consequently, how to better reconstruct 3D face models from 2D face images has become an extremely important and popular research topic.

The existing methods for 3D face reconstruction can be generally divided into 3 categories: 3D modeling methods based on images processing, pre-defined model based methods and end-to-end deep learning methods. The traditional 3D face reconstruction are mostly based on images features like brightness and edges. In comparison, model based 3D face reconstruction is currently getting more popular. In this method, 3D models are mainly represented by triangular meshes or point clouds. For example, the 3D Morphable Model\cite{3DMM} and CANDIDE-3\cite{CANDIDE} are the one of the most common used 3D face models in research field.

However, the above-mentioned methods are faced with some challenges. For instance, facial landmarks may become invisible due to self-occlusion when faces deviate from the frontal view. Additionally, traditional nonlinear models or cascaded linear regression are not complicated enough to cover all the facial features.

In recent years, some works started to apply deep learning network to estimate the 3DMM parameters of face models. However, the performance is restricted due to the limitation of the pre-defined face model. To avoid the limitation from certain face models, end-to-end 3D face reconstruction methods\cite{3DMM-CNN1,3DMM-CNN2,3DMM-CNN3} are proposed to get their own face models representation in the process without using pre-defined models.

\section{Related Work}
\label{RW}
3D Morphable Model (3DMM) is a widely used parameterized face model in 3D face reconstruction. In 3DMM, model features like facial shape, texture, camera position and light intensity are all determined by its coefficients. By calculating these coefficients, new face model can be automatically registered to a pre-defined face template\cite{3DMM}. Early methods tend to first establish the correspondences of the special points (feature points\cite{3DMM-local-feature} and landmarks\cite{landmarks}) between input images and the output 3D face model. Then, the non-linear optimization function is solved to regress the 3DMM coefficients. However, these methods highly rely on feature or landmark detectors which are often unstable. In recent years, some literature start to use Convolutional Neural Networks (CNNs) to estimate the 3DMM coefficients\cite{3DMM-CNN1,3DMM-CNN2,3DMM-CNN3}, which simplifies the process of calculation. However, the performance of 3DMM-based methods is restricted due to the limitation of the pre-defined face models\cite{Feng_2018_ECCV}.

To address the aforementioned issues, some literature start to build end-to-end models that take still 2D image as input and output the 3D face model directly. In \cite{jackson2017vrn}, the authors designed the volumetric representation of the 3D facial geometry, based on which the Volumetric Regression Network (VRN) is proposed to make spatial predictions at a voxel level. This network accepts an RGB image as input and directly regresses a 3D volume which can converted to a 3D face model. It has been proven that the proposed VRN outperforms all of the 3DMM-based methods and is able to deal with facial expressions as well as occlusions.
But the VRN outputs a volume with the dimensions of 192x192x200, which severely limits its computational efficiency. In \cite{Feng_2018_ECCV}, a novel UV position map is designed to records the 3D shape of human face in UV space. Additionally, in order to keep the semantic meaning of points within UV position map, 3DMM-based UV coordinates are created. Afterwards, Position map Regression Network (PRN) is proposed to regress the 3D full structure. This network takes 2D facial image as well as the ground-truth 3D shapes as input and directly outputs the corresponding 3D face model.

Although both VRN and PRN achieve state-of-the-art performance in 3D face reconstruction, the resolution of reconstructed model largely depends on the size of the feature map. Therefore, in order to obtain a high-resolution face model, a large feature map has to be used, resulting in high memory cost and low speed.

Compared with the aforementioned dense vertices methods which are typically based on fully connected network, the regression of 3DMM parameters with low dimension and low redundancy usually consumes far less time and memory. Starting from this point, Guo, X. Zhu, Y. Yang, F. Yang, Z. Lei, and S. Z. Li proposed 3DDFA-V2\cite{3DDFA-V2} to regress the parameters of 3DMM with a lightweight network. In the 3DDFA-V2 architecture, a meta-joint optimization strategy is utilized to dynamically regress a small set of 3DMM parameters which greatly enhances both speed and accuracy. More importantly, the core work of the novel 3DDFA-V2 method is the 3D aided short video synthesis which can simulate the face movement in and out of the plane and convert a still image into short video to improve the stability. Additionally, the proposed 3DDFA-V2 also impose an extra landmark-regression regularization to facilitate the parameters regression. With these three newly designed components, the 3DDFA-V2 has achieved high accuracy, fast speed and stability simultaneously, which is a milestone in the field of 3D face reconstruction and face alignment.

\section{Literature Review}

\subsection{Eye tracking and animation for MPEG-4 coding}
In 1998, N. Rossol, I. Cheng, W. F. Bischof, and A. Basu proposed an improved facial feature detection and tracking algorithms based on simple heuristics, and used the algorithm to build a model that presented the captured eye movement.\cite{eye}. The proposed approach first uses Hough transformation and deformable template matching, as well as color information to extract the eyes from a set of images sequence. Then, these extracted eye features are used to reconstruct the eye motions on a 3D facial model. From the experiment results, the proposed algorithms successfully synthesized high-quality eye movements, but the time complexity also increased significantly. In the future, the extensions of rebuilding lip movements and network strategies for real-time modelling will be discussed.

\subsection{QoE-Based Multi-Exposure Fusion in Hierarchical Multivariate Gaussian CRF}
In the paper \emph{QoE-Based Multi-Exposure Fusion in Hierarchical Multivariate Gaussian CRF}\cite{QoE}, R. Shen, I. Cheng, and A. Basu proposed a novel image fusion algorithm based on perceptual quality measures. To specify, perceived local contrast and color saturation are the measures used in this paper. Generally, perceived local contrast stands for the perception of local luminance variations of the surrounding luminance. However, there are many methods to measure local contrast. In order to deal with the noise in under-exposed regions, a threshold is introduced so that the under-exposed area will be brighter and preserves more details while the over-exposed area will be less vivid. Nevertheless, local contrast measure only works in the luminance channel so we need to use color saturation to supply it, especially when objects are captured at proper exposures and they have already shown saturated colors.

According to the experiment result, the fusion algorithm based on perceptual quality measures outperforms other methods, but the exploration of potential perceptual quality measures remains an open issue. In the future, we can try other measures that can be used in multi-exposure fusion.

\subsection{A framework for adaptive training and games in virtual reality rehabilitation environments}
In 2011, N. Rossol, I. Cheng, W. F. Bischof, and A. Basu proposed a framework for adaptive training and games in virtual reality rehabilitation environments\cite{VRGame}. They built a customizable VR rehabilitation environment where clinicians are able to design, build, and customize their indoor training environments for the patients. Moreover, since Bayesian network can determine patient's skill level, the environment can dynamically respond to user actions and change itself correspondingly. According to the experiment result, participants who used the proposed VR system accomplished the real world obstacle course quicklier, with a mean time of 81.5 seconds while it is 104.5 seconds for the control group. 

However, the proposed system was tested on non-disabled individuals, so further experiments need to be done to obtain a conclusive result. Besides, some research can be done on larger patient group to identify the potential problem of the system.

\subsection{Perceptually Guided Fast Compression of 3-D Motion Capture Data}
In the same year, A. Firouzmanesh, I. Cheng, and A. Basu proposed a fast compression of motion capture data technique\cite{compression}. In this method, wavelet coding is selected as the core algorithm due to its low complexity, high compression ratios and prone to adjust the perceptual quality of motion. Experiments show that the proposed method is much faster than other comparable algorithms, and is suitable to be used in mobile devices. However, in its wavelet algorithm, only bone lengths and variation in rotation are picked as factors to optimize the coefficient selection algorithm, while there are many other factors for us to explore.

\subsection{Panoramic video with predictive windows for telepresence applications}
Telepresence applications are widely used in various fields, such as explosives disposal telemedicine and so on\cite{telepresence}. To create an effective telepresence application, it is important to transmit the panoramic video effectively.
The paper proposed a predictive Kalman filter which can be applied to predict viewing direction for displaying panoramic images.
From the experiment, the prediction method and panoramic video are helpful to enhance the operator interface for the telepresence system.

\subsection{Gaussian and Laplacian of Gaussian weighting functions for robust feature based tracking}
The performance of object tracking algorithms is mainly limited by noise in environment, shape characteristic of objects an so on. Weighting functions are used to assign weights to different pixels in the images so that the object tracking system will can emphasize the important features in images. 

The paper\cite{Gaussian} proposed Gaussian and Laplacian of Gaussian weighting functions to improve the performance of the KLT tracking algorithm.

From the result of experiment, the proposed Gaussian weighting function and LOG weighting function both outperform the averaging weighting function of KLT when tracking objects in random noise environment, although the computation consumption increase by 10\%. Besides, it s also found that LOG weighting function works better at images with clear and sharp corners while Gaussian weighting function is better at dealing with real images which typically do not have sharp edges.

\subsection{Nose shape estimation and tracking for model-based coding}
In the traditional facial recognition system, eyes and mouth are always considered to be the most important features to detect. But actually the feature of nose shape is of the same importance to track and detect when analyzing or synthesising human facial expression.

The paper\cite{Nose} proposes a facial feature detection method which first put facial features into corresponding windows by using the global region growing and local region growing method. Then is to extract the accurate shapes of facial features and finally a tracking model can match with individual faces to track facial expression.

The experiments on real video sequences prove the facial feature detection method  for low bit rate video coding is applicable in practice.

\subsection{Hough transform for feature detection in panoramic images}
The omni-directional sensor can be utilized to obtain a 360 degrees field of view, which can be achieved simply by a camera through a radially symmetrical mirror and a traditional lens system. However, most existing mirror profiles violate the Single View Point (SVP) standard. Therefore, many existing methods cannot guarantee that the function of the lens system is equivalent to the standard perspective projection, which is a huge challenge.
Such an imaging system with a non-SVP optical system does not benefit from the affine quality of straight line features represented as collinear points in the image plane. In order to use these non-SVP images, a new method is needed to detect these features. This paper\cite{Hough}proposed a panoramic non-SVP image feature detection method based on improved Hough transform. The mathematical model of the feature extraction process is given.

Experiments and results show that the panoramic Hough transform can effectively detect distorted horizontal lines to supplement the function of vertical line detection in panoramic images. This transformation greatly improves the detection characteristics of the catadioptric panoramic sensor with a non-SVP mirror profile, and has good robustness.

\subsection{Generating Realistic Facial Expressions with Wrinkles for Model-Based Coding}
3D face reconstruction has always been a hot topic in academic research. However, the current face reconstruction technology can only roughly imitate the real face. It is very difficult to generate real facial texture . To solve this problem, this paper\cite{Generating} proposes a facial expression synthesis method based on local texture update. Firstly, a color based deformable template matching method is used to estimate the reference points on the face. In addition, an extended dynamic mesh matching face tracking algorithm is proposed. Compared with the whole face texture compression scheme, this method can significantly reduce the computational complexity and update quality of the whole face texture.

Experimental results show that, compared with the overall texture updating method, the new strategy based on local wrinkle texture updating has some advantages, which greatly reduces the number of coefficients to be transmitted. In addition, the proposed method can also maintain the high fidelity of image quality.

\subsection{Stereo Matching Using Random Walks}
Binocular stereo matching has always been a research hotspot of binocular vision. Binocular camera takes left and right view images of the same scene, uses stereo matching algorithm to obtain disparity map, and then obtains depth map. The range of application of depth map is very wide, because it can record the distance between the object in the scene and the camera, it can be used for measurement, 3D reconstruction, and virtual view synthesis.

This paper\cite{Stereo} proposes a two-phase stereo matching algorithm based on random walk frame. Random walks has been widely used in many fields. The main idea of random walk algorithm is to modeled the real-world problem as a muti-dimension graphic problem. From calculating the probability of randomly walking from one node to another, we can get the weights from considering some local quality globally.

In this proposed method, a set of reliable pixel matching algorithms based on Laplacian matrix and neighborhood information. Then, using reliable set as seed, the difference of unreliable regions is determined by solving Dirichlet problem. In order to improve the accuracy of parallax map, the change of illumination between different images is considered when building prior matrix and Laplacian matrix.

\subsection{Motion detection using background constraints}
Motion detection has become a hot research topic and can be applied in many fields. For example, autonomous driving requires the function of motion detection, which combines the research of motion planning and motion detection. In order to avoid collisions, vision-based navigation systems need to recognize the presence of stationary and moving objects.
This paper\cite{Motion} proposes a plane motion detection method based on background constraints. The camera is mounted on a platform that can rotate and translate. Using the information about the camera movement, the corresponding mapping of the pixels in the continuous image can be obtained to detect the movement.

The basic idea of this method is that if any point in the image is static, the point should satisfy the background constraint, while the point on an independent moving object is unlikely to satisfy the constraint. Therefore, it becomes easy to detect moving objects and can be used for translation and rotation.

Experiments and results show that the proposed motion detection method based on background constraints has strong robustness.

\subsection{Modeling Fish-eye Lenses}
A fish-eye lens is generally composed of a dozen different lenses. During the imaging process, the incident light is refracted to different degrees and projected onto an imaging plane with a limited size, making the fish-eye lens compared with ordinary lenses. 

Compared with ordinary pinhole camera, fish eye camera can observe a wider range, which can enhance the visual robustness to a certain extent. This paper\cite{Modeling} proposed a simple transformation that can be used to model variable resolution images. The experimental results show that our model works very closely with the real fish-eye lens.

\subsection{Event Dynamics Based Temporal Registration}
The video consists of a series of static images. However, when the video shakes severely, the image information in the video will become blurred. At this time, time registration can be used to make the video image more stable, which has been widely used in military, medical and other fields. In order to improve the effect of traditional Temporal Registration, the paper\cite{Event} proposed the Event Dynamics Based Temporal Registration method, and verified its effectiveness with experiments.

\subsection{Pose Recognition using the Radon Transform}
Human posture recognition technology is to recognize posture actions by collecting all kinds of physical information generated by human movement, including body posture, finger movement, facial movement and so on.

The research on the analysis and recognition of human motion and posture has become a hot direction, and has been applied in various fields such as medical rehabilitation, safe driving and robot technology. The existing human pose recognition methods can be roughly divided into model-based and non model-based.

This paper\cite{Pose} proposed a method to recognize human postures and gestures by using radon transformation. Radon transform is an integral transform, which projects the image space into the parameter space of the line in the form of line integral.

\subsection{Subjective and Objective Visual Quality Assessment of Textured 3D Meshes}
With the development of 3D printing and true 3D display technology, 3D grid related technology has become a research hot spot again. With the rapid development of 3D modeling technology, 3D models and their applications have received extensive attention. As the main form of three-dimensional model, three-dimensional grid model is widely used in computer-aided design, medical imaging, digital entertainment and other fields. However, the 3D mesh model is often distorted to different degrees after processing, which makes the visual quality of the 3D mesh model change. Therefore, it is very important to find an effective perceptual metric to evaluate the quality of these texture model artifacts.

This paper \cite{Subjective} designed and established a new subjective evaluation database texture 3D grid. In addition, the paper\cite{Subjective} also proposed a new objective index for the evaluation of the visual quality of texture grids as an optimized linear combination of grid quality and texture quality.

\section{Conclusion}
\label{conclusion}

3D face reconstruction and face alignment are the basis of real-time face animation generation. After our research on the related work, we identified that PRN provides us with an end-to-end method to solve these two problems simultaneously. In this project, we will try to add data augmentation as well as the post-processing SFS method to improve the robustness and the precision of PRN. Besides, we will also explore other potential CNN architectures to further enhance the performance of our model.

\section{References}


\bibliographystyle{IEEEtran}
\bibliography{egbib}

\end{document}